\documentclass[conference]{IEEEtran}
\IEEEoverridecommandlockouts

\usepackage{amsmath,amssymb,amsfonts}
\usepackage{booktabs}
\usepackage{multirow}
\usepackage{colortbl}
\usepackage{array}
\usepackage{multicol}
\usepackage{hyperref}

\usepackage{algorithm}
\usepackage{algpseudocode}

\usepackage{graphicx}
\usepackage{tikz}
\usepackage{caption}
\usepackage{subcaption}
\usepackage{xcolor}
\usepackage{colortbl} 
\usepackage{orcidlink}

\def\BibTeX{{\rm B\kern-.05em{\sc i\kern-.025em b}\kern-.08em
    T\kern-.1667em\lower.7ex\hbox{E}\kern-.125emX}}
\begin{document}

\title{Point-RTD: Replaced Token Denoising for Pretraining Transformer Models on Point Clouds\\
\thanks{This work was supported in part by the National Science Foundation under Grant No. OIA-2148788.}

}

\author{
\IEEEauthorblockN{Gunner Stone\orcidlink{0000-0002-5182-1365}, Youngsook Choi\orcidlink{0009-0007-9642-4666},Alireza Tavakkoli\orcidlink{0000-0001-9460-1269}, and Ankita Shukla\orcidlink{0000-0002-1878-2667}}
\IEEEauthorblockA{\textit{Department of Computer Science and Engineering, University of Nevada, Reno,}
USA}
}

\maketitle

\begin{abstract}
Pre-training strategies play a critical role in advancing the performance of transformer-based models for 3D point cloud tasks. In this paper, we introduce Point-RTD (Replaced Token Denoising), a novel pretraining strategy designed to improve token robustness through a corruption-reconstruction framework. Unlike traditional mask-based reconstruction tasks that hide data segments for later prediction, Point-RTD corrupts point cloud tokens and leverages a discriminator-generator architecture for denoising. This shift enables more effective learning of structural priors and significantly enhances model performance and efficiency. On the ShapeNet dataset, Point-RTD reduces reconstruction error by over 93\% compared to PointMAE, and achieves more than 14× lower Chamfer Distance on the test set. Our method also converges faster and yields higher classification accuracy on ShapeNet, ModelNet10, and ModelNet40 benchmarks, clearly outperforming the baseline Point-MAE framework in every case. Code is available at \url{https://github.com/GunnerStone/PointRTD}. \end{abstract}

\section{Introduction}
Point clouds have become an essential data representation in various fields such as remote sensing, autonomous driving, and robotics \cite{guo2020deep}. They provide a rich three-dimensional description of the environment, enabling critical tasks like object detection\cite{misra2021end}, classification and segmentation \cite{qi2017pointnet} to name a few. However, unlike structured data like text sequences or image grids, point clouds are inherently unstructured and lack both intrinsic ordering and uniform neighborhood relationships. This lack of intrinsic ordering and uniformity complicates the application of transformer-based architectures that have seen great success in NLP~\cite{vaswani2017attention} and computer vision \cite{dosovitskiy2020image}. 

In the realm of point cloud processing, various strategies have been developed to adapt transformer architectures to better handle this unstructured data. These include point-based methods, where each point is treated as an individual token~\cite{zhao2021point, wu2022point}, voxel-based approaches that convert point clouds into voxel grids to utilize volumetric features~\cite{mao2021voxel}, and dual modality strategies that attempt to leverage both point and geometry-preserved 2D-projections~\cite{wang2022p2p}. Despite the variety of approaches, the most widely adopted models employ patch-based tokenization, that segment point clouds into clusters or patches. This category includes several prominent models such as Point-BERT~\cite{yu2022point}, Point-MAE~\cite{pang2022masked}, among others~\cite{zhang2022point, zha2024towards, qi2023contrast, qi2024shapellmuniversal3dobject, guo2021pct}.

While these popular patch-based models predominantly employ a masked autoencoding pretraining regime, where portions of the data are obscured and the model is trained to predict these missing parts, this approach has not been definitively established as the optimal strategy for reconstructing point cloud data. Masked autoencoding provides a foundational technique for learning robust features, yet there may be alternative strategies that could offer enhanced learning capabilities in dealing with the unique challenges of unstructured 3D spaces.

To explore these possibilities, we introduce Point-RTD (Replaced Token Denoising), which extends the concept of replaced token denoising - originally inspired by GANLM~\cite{yang2023ganlmencoderdecoderpretrainingauxiliary} - to the realm of point clouds. This approach enhances token robustness and reinforces semantic consistency through a structured corruption-reconstruction framework, representing a notable alternative to traditional latent-space embedding techniques.

\begin{figure}[t]
    \centering
    \includegraphics[width=0.85\linewidth]{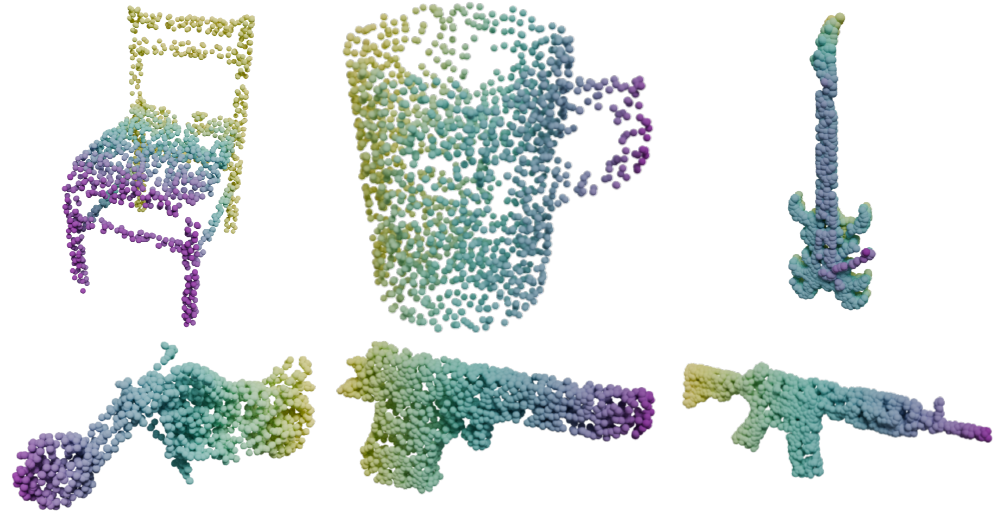}
    \caption{
        Example reconstructions from test-split of ShapeNet using our PointRTD model. Each reconstruction achieves a Chamfer Distance below 0.221\(\times 10^{-3}\).
    }
    \label{fig:shapenet_reconstructions}
\end{figure}

\textbf{Our Contribution:}
We propose Point-RTD, a novel pretraining strategy for point cloud transformers based on replaced token denoising, introducing a new perspective for tokenization and self-supervised learning in 3D data.

Our approach Point-RTD is evaluated across several well-known point cloud benchmark datasets to help demonstrate its advantages in terms of robustness, efficiency, and accuracy. On ShapeNet, Point-RTD reduces reconstruction error (Chamfer Distance) by over 93\% on the test set compared to Point-MAE, achieving 0.221\(\times 10^{-3}\) vs. 2.81\(\times 10^{-3}\), as illustrated by the high-fidelity reconstructions in Figure~\ref{fig:shapenet_reconstructions}. It also converges faster on Modelnet10 and delivers higher classification accuracy on Modelnet40, highlighting the effectiveness of our corruption-reconstruction framework over traditional masking-based strategies.

\begin{figure*}[ht!]
    \centering
    \includegraphics[width=\textwidth]{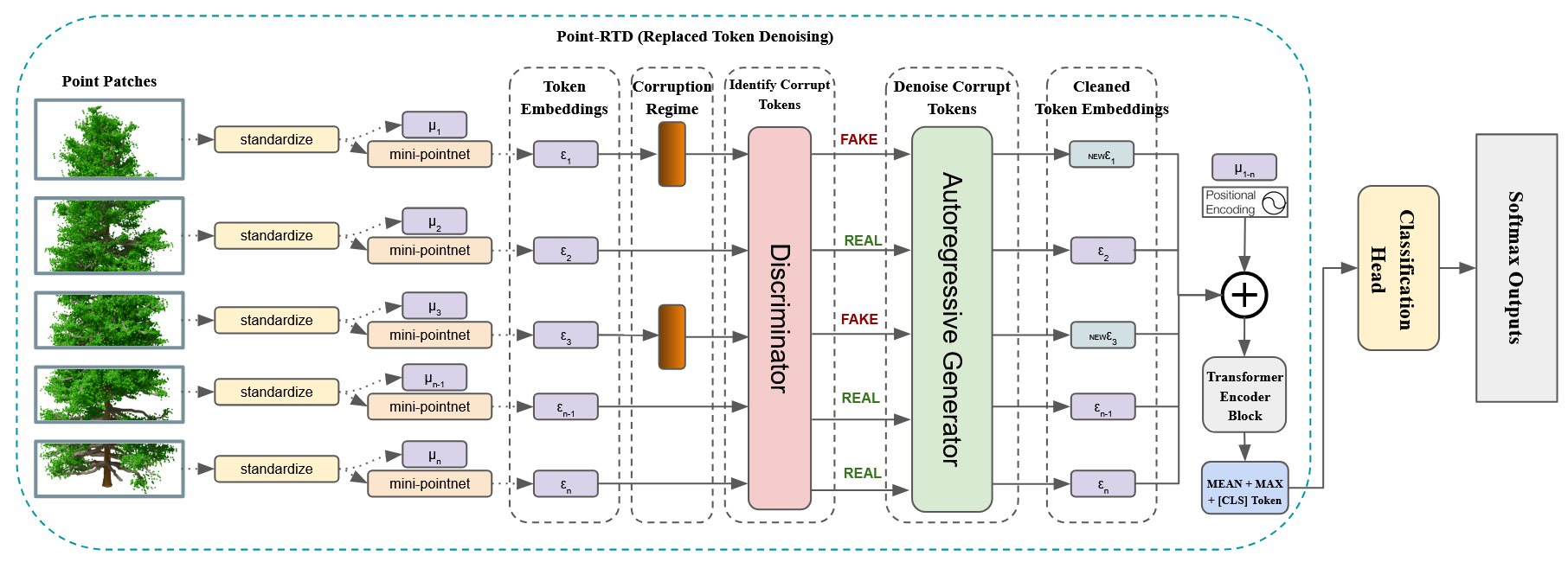}
    \caption{Overview of the Point-RTD architecture. Point patches are tokenized using FPS+KNN and a mini-PointNet. A subset of tokens is corrupted, then passed through a discriminator to identify them. The generator denoises the flagged tokens, while clean ones remain unchanged. The resulting tokens are encoded by Transformer blocks and used for downstream tasks.}
    \label{fig:architecture}
\end{figure*}

\section{Methodology}
An overview of proposed approach Point-RTD is shown in Figure \ref{fig:architecture}.  
\subsection{Point Tokenization and Preprocessing}
Point-RTD utilizes patch-based tokenization, similar to Point-BERT~\cite{yu2022point} and Point-MAE~\cite{pang2022masked}, segmenting point clouds into patches using Farthest Point Sampling (FPS) and k-Nearest Neighbors (kNN). Input patches are normalized by subtracting their centroid from each point in the patch before being processed through a mini-PointNet~\cite{qi2017pointnetdeeplearningpoint}, which encodes them into token embeddings that capture local geometric features.

\subsection{Corruption Regime}
To encourage the model to learn robust representations, we apply corruption to token sequences during training. Initially, this was done by applying multiplicative Gaussian noise \(\mathcal{N}(0, 1)\) to a large percentage (80\%) of the tokens. This noise disrupts the token's original feature distribution and forces the model to rely on the remaining uncorrupted tokens to reconstruct the input. While this approach serves as a form of regularization by encouraging the inference of missing information, it primarily operates at the individual token level and does not fully leverage inter-sample or inter-class relationships.

To further enhance the regularization effect, we extend the corruption regime by introducing a \textit{token replacement} strategy. Instead of applying noise, a subset of tokens is replaced with tokens from other samples in the batch. This can be done in two ways:

\begin{itemize}
    \item \textbf{Random mixup}: Corrupted tokens are replaced with randomly selected tokens taken from a different sample within the batch.
    \item \textbf{Nearest-neighbor mixup}: Corrupted tokens are replaced with tokens from the most similar sample of a different class, based on the distance between average token embeddings.
\end{itemize}

This replacement-based corruption can be viewed as a form of \textit{hard token mixup}, which challenges the model with semantically mixed inputs. It acts as a stronger regularizer by enforcing the model to learn class-distinctive representations that remain robust under class-mismatched tokens.

While nearest-neighbor replacement can be beneficial in disambiguating semantically close classes (e.g., \textit{plant} vs.\ \textit{vase}), it may overly specialize on such boundaries, potentially limiting generalization. In contrast, random token replacement introduces greater diversity in corruption patterns, exposing the model to a broader range of inter-class interactions. Empirically, we observe that random mixup yields better performance, indicating that diverse and unconstrained corruption is more effective at regularizing the model's latent space.

Conceptually, we view the corruption regime in Point-RTD as a form of contrastive regularization in token space. By replacing meaningful local structures with semantically inconsistent alternatives, the model is implicitly trained to minimize confusion across class boundaries. This setup parallels principles in contrastive learning, where negative samples push representations apart. However, Point-RTD does so within the denoising objective, softening the need for explicit contrastive pairs or loss functions. Consequently, the model benefits from both reconstruction-driven learning and implicit semantic separation, leading to more discriminative and robust embeddings.

This corruption regime can be applied either at the token embedding level (before transformer processing) or at the contextual token level (after transformer layers). The former reflects localized features such as patch shapes, while the latter captures those shapes in relation to their surrounding context. Empirically, we find that applying corruption to contextual tokens yields slightly better performance, likely because it encourages robustness in more semantically meaningful and globally-informed representations.

\subsection{Discriminator and Generator}
The discriminator model identifies whether tokens are corrupted or clean. During training, the discriminator uses a weighted binary cross-entropy loss to distinguish real and fake tokens and manage balancing them to help control the training dynamics.

The generator autoregressively cleans corrupted tokens, guided by the discriminator’s feedback. It minimizes mean squared error (MSE) between the cleaned and original tokens, optimizing for accurate reconstruction of token embeddings.

\subsection{Pretraining Algorithm}
We evaluated the effectiveness of our proposed Point-RTD framework by pretraining it alongside Point-MAE on a reconstructive task using the ShapeNet dataset~\cite{shapenet2015}. ShapeNet consists of approximately 51,300 clean 3D models across 55 common object categories. Following the setup described in the Point-MAE paper~\cite{pang2022masked}, we split the dataset into training and validation sets, using only the training set for pretraining. The pretraining process, including the corruption, discrimination, generation, and reconstruction steps, is detailed in Algorithm~\ref{alg:rtd}.

\begin{algorithm}
\footnotesize
\caption{Replaced Token Denoising (RTD) for 3D Point Clouds}\label{alg:rtd}
\begin{algorithmic}[1]
\State \textbf{Input:} Point patches $\mathbf{P} = \{\mathbf{p}_1, \mathbf{p}_2, \dots, \mathbf{p}_n\}$, mini-PointNet encoder, discriminator $D$, generator $G$, Transformer encoder $\mathcal{T}$, and Transformer decoder $\mathcal{D}ecoder$
\State \textbf{Output:} Cleaned token embeddings $\hat{\mathbf{t}}$

\State \textbf{Initialize} Gaussian noise $\mathcal{N}(0, 1)$
\State \textbf{Tokenization:} $\mathbf{t}_i \gets \texttt{mini-PointNet}(\mathbf{p}_i)$
\State \textbf{Corruption:} $\tilde{\mathbf{t}}_i[j] \gets \mathbf{t}_{k}[j']$, where $k \neq i$, $y_k \neq y_i$, for 80\% of token indices $j$
\State \textbf{Discrimination:} $y_i \gets D(\tilde{\mathbf{t}}_i)$
\State \textbf{Generation:} $\hat{\mathbf{t}}_i \gets 
\begin{cases}
      \mathbf{t}_i, & \text{if } y_i = \text{REAL}, \\
      G(\tilde{\mathbf{t}}_i), & \text{if } y_i = \text{FAKE}.
\end{cases}$
\State \textbf{Encoding:} $\hat{\mathbf{c}} \gets \mathcal{T}(\hat{\mathbf{t}})$ with positional encodings
\State \textbf{Reconstruction:} $\hat{\mathbf{P}} \gets \mathcal{D}ecoder(\hat{\mathbf{c}})$
\State \textbf{Loss Functions:}

\[
\mathcal{L}_D = -w_{\text{real}} \cdot \mathbb{E}[\log D(\mathbf{t})] - w_{\text{fake}} \cdot \mathbb{E}[\log(1 - D(\tilde{\mathbf{t}}))]
\]
\[
\mathcal{L}_G = \|\hat{\mathbf{t}} - \mathbf{t}\|_2^2
\]
\[
\mathcal{L}_{\text{Chamfer}} = \text{ChamferDistance}(\hat{\mathbf{P}}, \mathbf{P})
\]
\State \textbf{Total Loss:} $\mathcal{L} \gets \mathcal{L}_D + \mathcal{L}_G + \mathcal{L}_{\text{Chamfer}}$
\State \textbf{Return} $\mathcal{L}$ for the pretraining objective and $\hat{\mathbf{t}}$ as feature embeddings for fine-tuning.
\end{algorithmic}
\end{algorithm}

\section{Experimental Methodology}
 In this section, we describe the implementation details and training strategies. 

\paragraph{Pretraining Details.} Both Point-RTD and Point-MAE models are pretrained for 150 epochs using the AdamW optimizer with a weight decay of 0.05 and a cosine learning rate decay scheduler. The initial learning rate was set to 0.001, with a batch size of 64. Point-MAE masked 80\% of the tokens, while Point-RTD corrupted 80\% of the tokens by applying random-mixup.

Using this shared pre-training framework, we asses the effectiveness of Point-RTD in learning robust and generalizable representations for downstream tasks compared to Point-MAE.

\paragraph{Fine-Tuning Experimental Setup}
Fine-tuning is performed using the ModelNet10 dataset~\cite{wu20153d}, comprising 4,899 CAD models covering 10 object categories. The preprocessing pipeline is consistent with that used during pre-training. We utilized the pretrained encoders from both Point-RTD and Point-MAE, adapting the learned features for classification tasks. Afterwards, we assessed the effectiveness of the representations learned during pretraining by evaluating the fine-tuned models' accuracy in classifying point clouds into their respective categories.

\paragraph{Encoder and Classification Head Setup}
Both Point-RTD and Point-MAE had their pretrained encoders connected to a unified classification head, as depicted in Fig \ref{fig:architecture}. The classification head aggregates encoder outputs by concatenating the \texttt{[CLS]} token, mean-pooled, and max-pooled token embeddings. These concatenated features are then processed through a single-layer fully connected network to produce class logits.

\paragraph{Finetuning Setup}
The models are trained for 300 epochs using the AdamW optimizer with an initial learning rate of \(10^{-3}\) and a weight decay of 0.05, applying random scaling and translation to input point clouds for improved generalization. Both models shared identical settings to ensure fair comparisons, using the provided train-test splits of the ModelNet10 dataset.

\section{Results and Analysis}

We evaluate Point-RTD against Point-MAE in both the pretraining phase and downstream classification task to assess reconstruction performance and feature generalization capabilities. For a fair comparison, we apply the same pretraining and fine-tuning settings as described in Section~3.

\begin{table}[h!]
    \centering
    \caption{Average Chamfer Distance \(\times 10^3\) between reconstructed and ground truth point clouds on ShapeNet train and test splits. Lower values indicate better reconstruction quality.}
    \label{tab:chamfer_distance}
    \resizebox{\columnwidth}{!}{%
    \begin{tabular}{lcc}
        \toprule
        \textbf{Pretrained Model} & \textbf{Train Set CD} & \textbf{Test Set CD} \\
        \midrule
        PointMAE & 1.046 & 2.805 \\
        Point-RTD (ours) & \textbf{0.208} & \textbf{0.221} \\
        \bottomrule
    \end{tabular}
    }
\end{table}

\begin{figure*}[t]
    \centering
    \includegraphics[width=.85\textwidth]{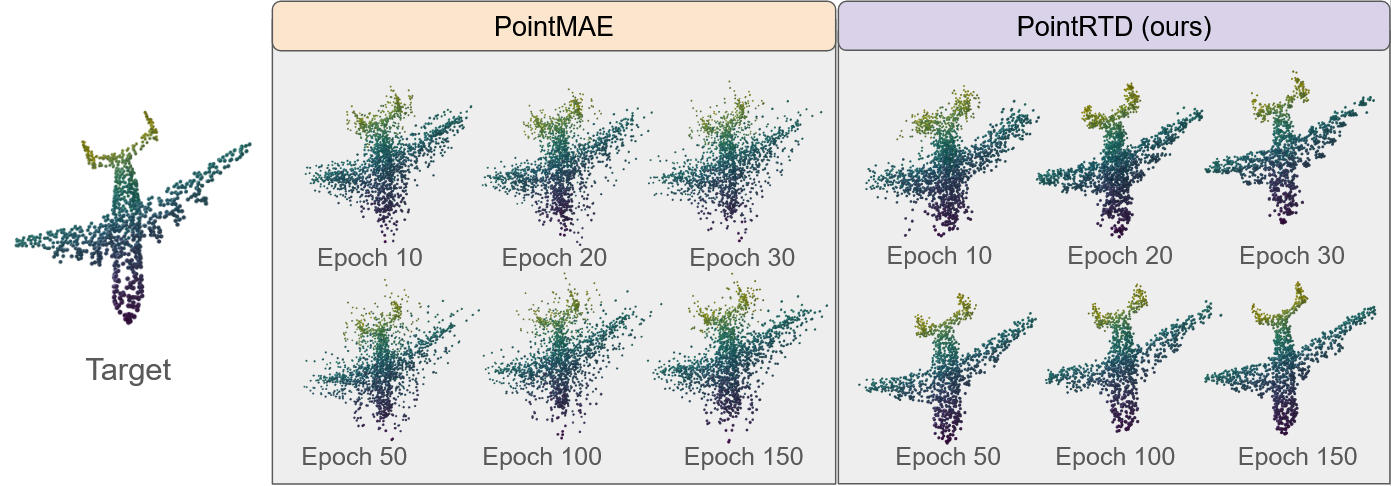}
    \caption{Qualitative reconstruction results from PointMAE and Point-RTD at various pretraining epochs. Point-RTD produces more coherent and complete structures across all stages, especially in early epochs, which aligns with its lower Chamfer Distance scores in Table~\ref{tab:chamfer_distance}.}
    \label{fig:pretraining_visual}
\end{figure*}

\begin{table}[t!]
\centering
\caption{\textbf{ModelNet10:} Classification accuracy (\%) at various training epochs. Models were evaluated using a 10-vote majority mechanism.}
\label{tab:results}
\begin{tabular}{@{}lcccccc@{}}
\toprule
 & \multicolumn{6}{c}{\textbf{Training Epochs (Checkpoints)}} \\
\cmidrule(l){2-7}
\textbf{Fine-tuned Model} & \textbf{300} & \textbf{250} & \textbf{200} & \textbf{150} & \textbf{100} & \textbf{50} \\
\midrule
PointMAE & 89.76 & 87.56 & 86.45 & 52.64 & 66.08 & 13.66 \\
Point-RTD (ours) & \textbf{92.73} & 92.29 & 91.96 & 90.20 & 86.89 & 87.22 \\
\bottomrule
\end{tabular}%
\end{table}
\vspace{-1.5mm}

\subsection{Pretraining Results}
The pretraining performance of both models is summarized in Table~\ref{tab:chamfer_distance}. Here, we use Chamfer Distance (CD) as the primary metric, scaled by \( \times 10^3 \) in the table for clarity. Point-RTD achieves an average CD of 0.208\(\times 10^{-3}\) on the training set and 0.221\(\times 10^{-3}\) on the test set, demonstrating an order of magnitude higher reconstruction fidelity compared to Point-MAE. In contrast, Point-MAE achieves a CD of 1.046 on the training set but exhibits a significantly higher CD of 2.805 on the test set, indicating potential generalization issues. Notably, Point-RTD maintains consistent reconstruction performance, with only an 8\% increase in error between the train and test sets, demonstrating significantly improved feature robustness and generalization to unseen data. These findings suggest that Point-RTD not only learns higher-quality features but also more transferable representations, making it a preferable choice for deployment.

To support the quantitative results, Figure~\ref{fig:pretraining_visual} provides a visual comparison of reconstructions produced by PointMAE and Point-RTD at different pretraining epochs. Point-RTD consistently generates more complete and structurally coherent outputs than PointMAE, even at earlier stages of training. These improvements in visual fidelity align with the Chamfer Distance trends shown in Table~\ref{tab:chamfer_distance}, reinforcing the effectiveness of our RTD regularization technique.

\subsection{Classification Results}
We assess the performance of Point-RTD and Point-MAE on the downstream task of ModelNet10 classification. The classification accuracy of both models was computed using a standard 10-vote majority mechanism, introduced by Liu et al.~\cite{liu2019relation}. During evaluation, due to the random point sampling inherent to the patch candidate process, each test sample was passed through the model 10 times, and the label receiving the most votes was chosen as the final prediction. This method, described in the Point-MAE paper for fair comparisons, ensures consistency in evaluation.

The results of the classification task, summarized in Table~\ref{tab:results}, demonstrate significant performance improvements achieved by the proposed Point-RTD framework over the baseline Point-MAE. Point-RTD achieves the highest classification accuracy of 92.73\%, compared to Point-MAE’s peak of 89.76\%. Moreover, Point-RTD exhibits substantially faster validation convergence, reaching 87.22\% accuracy after just 50 epochs, while Point-MAE lags at 13.66\%. This rapid convergence indicates a drastic difference in Point-RTD’s ability to learn robust token representations during pre-training more efficiently under identical training constraints. At 150 epochs, Point-RTD achieves 90.20\%, already surpassing Point-MAE’s final accuracy at 300 epochs of 89.76\%.

\subsection{Validating Representational Strength via ModelNet40}

\begin{table}[h!]
\centering
\small
\caption{\textbf{ModelNet40}: We report shape classification accuracy (\%) using linear SVM and fine-tuning with and without voting. Gray rows use cross-modal information or extra pretraining.}
\label{tab:modelnet40}
\begin{tabular}{@{}lccc@{}}
\toprule
\multirow{2}{*}{\textbf{Method}} & \multicolumn{3}{c}{\textbf{ModelNet40}} \\
\cmidrule(l){2-4}
 & \textbf{SVM} & \textbf{w/o Voting} & \textbf{w/ Voting} \\
\midrule
PointBERT~\cite{yu2022point}      & 87.4 & 92.7 & 93.2 \\
MaskPoint~\cite{liu2022masked}         & --   & -- & 93.8 \\
MAE3D~\cite{jiang2023masked}            & -- & 93.4 & --   \\
PointMAE~\cite{pang2022masked}       & 92.7 & 93.2 & 93.8 \\
PointGame~\cite{liu2023pointgame}      & 90.0 & 93.1 & 93.5 \\
PointMA2E~\cite{zhang2022point}     & 92.9   & 93.4 & 94.0 \\
PointGPT-S~\cite{chen2023pointgpt}     & --   & -- & 94.2 \\
\textbf{PointRTD (ours)}               & \textbf{93.0} & \textbf{94.1} & \textbf{94.2} \\
\midrule
\textcolor{gray}{CrossNet~\cite{wu2023self}} &
\textcolor{gray}{91.5} & \textcolor{gray}{93.4} & \textcolor{gray}{--} \\
\textcolor{gray}{InterMAE~\cite{liu2023inter}}        &
\textcolor{gray}{--} & \textcolor{gray}{93.6} & \textcolor{gray}{--} \\
\textcolor{gray}{ACT~\cite{dong2022autoencoders}}               &
\textcolor{gray}{--} & \textcolor{gray}{93.7} & \textcolor{gray}{94.0} \\
\textcolor{gray}{I2P-MAE~\cite{zhang2023learning}}       &
\textcolor{gray}{--} & \textcolor{gray}{93.7} & \textcolor{gray}{94.1} \\
\textcolor{gray}{ReCon~\cite{qi2023contrast}}          &
\textcolor{gray}{--} & \textcolor{gray}{94.5} & \textcolor{gray}{94.7} \\
\textcolor{gray}{PointGPT-L~\cite{chen2023pointgpt}}  &
\textcolor{gray}{--} & \textcolor{gray}{--} & \textcolor{gray}{94.7} \\
\bottomrule
\end{tabular}%
\end{table}

While the primary motivation behind Point-RTD is its training efficiency and strong performance on downstream tasks like ModelNet10, we also demonstrate that the method is competitive with other powerful point cloud models on the widely adopted ModelNet40 benchmark. This section provides further insight that complements earlier results by showing that Point-RTD not only converges more rapidly to a robust latent embedding space, but also maintains high accuracy when scaled to more challenging datasets.

To assess the broader competitiveness of our approach, we trained a Point-RTD model using a high corruption ratio of 80\%. Corruption was applied using a random mixup strategy, where corrupted tokens were replaced with tokens from other samples within the same batch. The model operated on input samples that were tokenized into 64 patches, each consisting of 32 points. Optimization was performed with the AdamW optimizer, using a learning rate of 0.001, cosine annealing schedule, and a weight decay of 0.05. 

\begin{figure}[h!]
    \centering
    \includegraphics[width=\columnwidth]{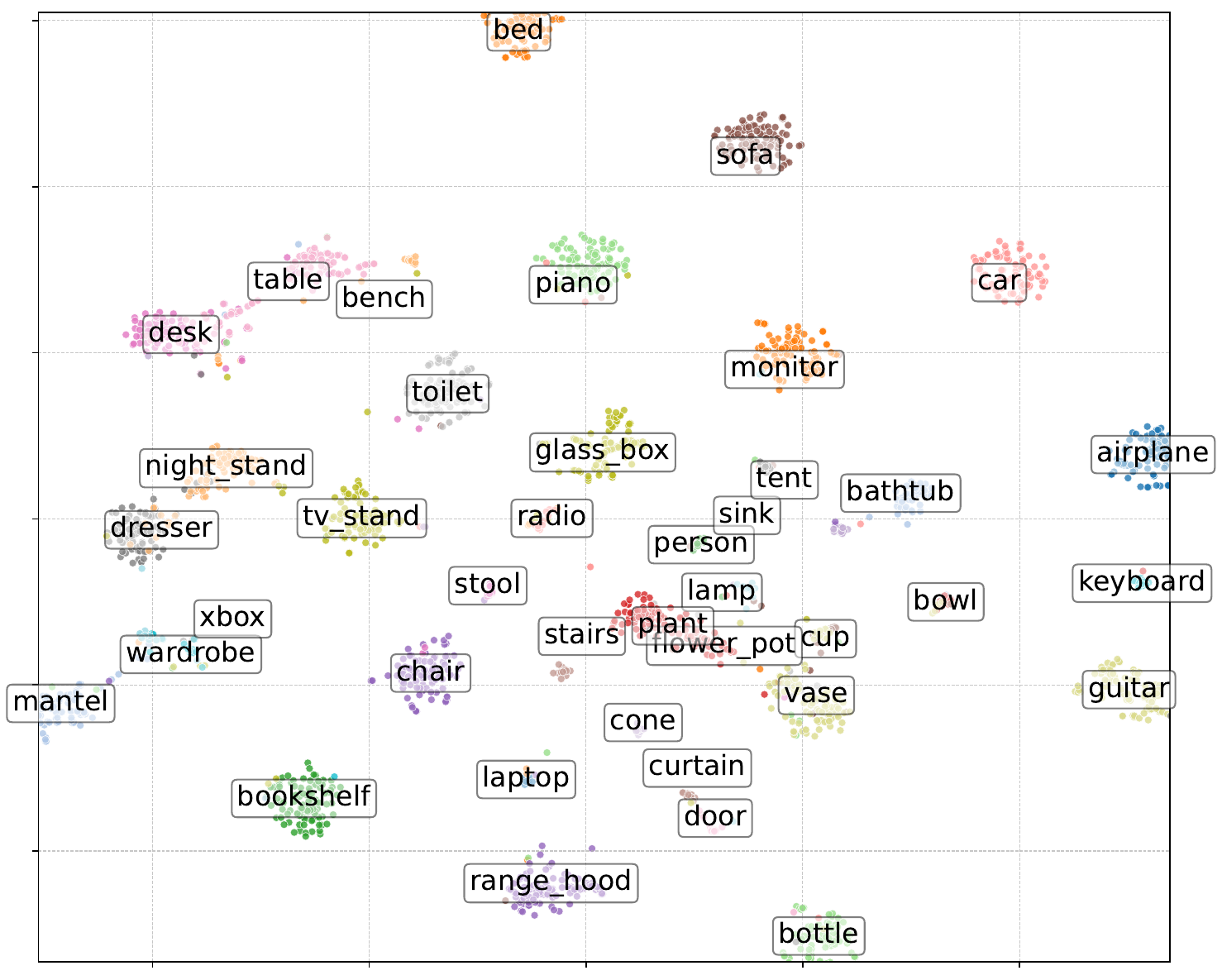}
    \caption{t-SNE visualization of the learned feature embeddings on ModelNet40 test set using Point-RTD. Each point represents a test set sample embedding, colored by class.}
    \label{fig:tsne_modelnet40}
\end{figure}

As shown in Table~\ref{tab:modelnet40}, Point-RTD achieves 94.2\% accuracy with a standard 10-vote majority mechanism on ModelNet40, surpassing or matching the performance of several strong baselines. Notably, our method also maintains strong linear SVM accuracy (93.0\%), indicating that the learned representations are linearly separable and well-structured. These results reinforce that Point-RTD is not only a fast learner but also highly competitive in accuracy compared to other methods, including those leveraging cross-modal information or additional pretraining.

Figure~\ref{fig:tsne_modelnet40} shows a t-SNE visualization \cite{van2008visualizing} of the learned feature embeddings from our model on the ModelNet40 test set. Each point represents an object embedding, colored by its ground-truth class. Interestingly, we observe that certain classes that are commonly confused in classification tasks, such as desk and table, dresser and nightstand, plant and flower pot, or cup and vase, tend to appear close together in the embedding space. This reflects well-known issues in the ModelNet40 dataset, where different object classes can exhibit nearly identical geometric structure. Prior studies \cite{mesika2021cloudwalker} have noted that misclassifications between such pairs are not necessarily indicative of model error, but instead result from ambiguous or inconsistent labeling in the dataset itself. For example, a flower pot in the training set might be geometrically indistinguishable from a vase in the test set. In such cases, the model may classify the object differently from its label, even though its prediction is arguably reasonable. The t-SNE plot provides a visual explanation of this phenomenon by revealing which classes tend to share similar embedding spaces.

\subsection{Discussion}

Our experiments show that Point-RTD outperforms Point-MAE on ModelNet10, using the same pretraining and fine-tuning hyperparameters reported as optimal in the original Point-MAE paper. With just 50 pretraining epochs, Point-RTD achieves 92.73\% test accuracy, demonstrating superior efficiency. Additionally, it converges significantly faster during fine-tuning, reaching 87.22\% accuracy in 50 epochs, whereas Point-MAE achieves only 13.66\% in the same period.

These findings highlight the broader implications of designing efficient pretraining frameworks. The strong performance of Point-RTD under identical training regimes challenges the idea that long pretraining schedules are necessary for high downstream accuracy. Instead, our method illustrates that robustness-centered pretraining can produce strong representations with significantly reduced computational effort.

The performance improvements observed with Point-RTD suggest that replaced token denoising imposes a stronger regularization signal than traditional masked autoencoding. By injecting semantically incorrect or cross-class tokens into the input, the model is forced to disambiguate real versus corrupted context, encouraging sharper inter-class boundaries and more generalizable features. This is especially beneficial for unstructured 3D data, where local geometry alone may be insufficient to distinguish semantically similar categories. The explicit discriminator-guided feedback loop further reinforces learning by highlighting areas of semantic mismatch. Compared to masked token recovery, which passively reconstructs withheld input, Point-RTD actively challenges the model’s understanding of object structure and semantics.

To further evaluate the competitiveness of Point-RTD, we conducted experiments on the more widely used ModelNet40 benchmark. Our method achieves 94.2\% accuracy with a 10-vote inference scheme, matching or outperforming several recent models while maintaining strong linear SVM performance (93.0\%). These results confirm that Point-RTD is not only efficient, but also competitive in absolute accuracy on standard benchmarks.

\section{Conclusion}

We introduced Point-RTD, a novel pretraining framework for point cloud transformers based on a corruption-reconstruction paradigm inspired by Replaced Token Denoising. By corrupting tokens and leveraging a generator-discriminator architecture for denoising, Point-RTD learns meaningful latent token embeddings for point cloud reconstruction, while also regularizing the latent space to improve inter-class feature separability for downstream classification tasks.

In this study, we focus on applying Point-RTD within the Point-MAE framework due to its foundational status and continued influence in the field. Point-MAE serves as a representative and well-established benchmark, allowing us to clearly demonstrate the benefits of our approach. However, the design of Point-RTD is model-agnostic: its corruption and denoising strategy is broadly applicable to any patch-based point cloud transformer. Given its architecture-agnostic design, the framework is well-suited for future extensions and adaptations.

Point-RTD provides an effective means of regularizing transformer-based models through pretraining, while also supporting strong performance in end-to-end training regimes. Its versatility allows it to be integrated into a wide variety of 3D vision pipelines.

\bibliographystyle{IEEEtran}
\bibliography{custom}

\begin{thebibliography}{10}
\providecommand{\url}[1]{#1}
\csname url@samestyle\endcsname
\providecommand{\newblock}{\relax}
\providecommand{\bibinfo}[2]{#2}
\providecommand{\BIBentrySTDinterwordspacing}{\spaceskip=0pt\relax}
\providecommand{\BIBentryALTinterwordstretchfactor}{4}
\providecommand{\BIBentryALTinterwordspacing}{\spaceskip=\fontdimen2\font plus
\BIBentryALTinterwordstretchfactor\fontdimen3\font minus \fontdimen4\font\relax}
\providecommand{\BIBforeignlanguage}[2]{{%
\expandafter\ifx\csname l@#1\endcsname\relax
\typeout{** WARNING: IEEEtran.bst: No hyphenation pattern has been}%
\typeout{** loaded for the language `#1'. Using the pattern for}%
\typeout{** the default language instead.}%
\else
\language=\csname l@#1\endcsname
\fi
#2}}
\providecommand{\BIBdecl}{\relax}
\BIBdecl

\bibitem{guo2020deep}
Y.~Guo, H.~Wang, Q.~Hu, H.~Liu, L.~Liu, and M.~Bennamoun, ``Deep learning for 3d point clouds: A survey,'' \emph{IEEE transactions on pattern analysis and machine intelligence}, vol.~43, no.~12, pp. 4338--4364, 2020.

\bibitem{misra2021end}
I.~Misra, R.~Girdhar, and A.~Joulin, ``An end-to-end transformer model for 3d object detection,'' in \emph{Proceedings of the IEEE/CVF international conference on computer vision}, 2021, pp. 2906--2917.

\bibitem{qi2017pointnet}
C.~R. Qi, H.~Su, K.~Mo, and L.~J. Guibas, ``Pointnet: Deep learning on point sets for 3d classification and segmentation,'' in \emph{Proceedings of the IEEE conference on computer vision and pattern recognition}, 2017, pp. 652--660.

\bibitem{vaswani2017attention}
A.~Vaswani, ``Attention is all you need,'' \emph{Advances in Neural Information Processing Systems}, 2017.

\bibitem{dosovitskiy2020image}
A.~Dosovitskiy, ``An image is worth 16x16 words: Transformers for image recognition at scale,'' \emph{arXiv preprint arXiv:2010.11929}, 2020.

\bibitem{zhao2021point}
H.~Zhao, L.~Jiang, J.~Jia, P.~H. Torr, and V.~Koltun, ``Point transformer,'' in \emph{Proceedings of the IEEE/CVF international conference on computer vision}, 2021, pp. 16\,259--16\,268.

\bibitem{wu2022point}
X.~Wu, Y.~Lao, L.~Jiang, X.~Liu, and H.~Zhao, ``Point transformer v2: Grouped vector attention and partition-based pooling,'' \emph{Advances in Neural Information Processing Systems}, vol.~35, pp. 33\,330--33\,342, 2022.

\bibitem{mao2021voxel}
J.~Mao, Y.~Xue, M.~Niu, H.~Bai, J.~Feng, X.~Liang, H.~Xu, and C.~Xu, ``Voxel transformer for 3d object detection,'' in \emph{Proceedings of the IEEE/CVF international conference on computer vision}, 2021, pp. 3164--3173.

\bibitem{wang2022p2p}
Z.~Wang, X.~Yu, Y.~Rao, J.~Zhou, and J.~Lu, ``P2p: Tuning pre-trained image models for point cloud analysis with point-to-pixel prompting,'' \emph{Advances in neural information processing systems}, vol.~35, pp. 14\,388--14\,402, 2022.

\bibitem{yu2022point}
X.~Yu, L.~Tang, Y.~Rao, T.~Huang, J.~Zhou, and J.~Lu, ``Point-bert: Pre-training 3d point cloud transformers with masked point modeling,'' in \emph{Proceedings of the IEEE/CVF conference on computer vision and pattern recognition}, 2022, pp. 19\,313--19\,322.

\bibitem{pang2022masked}
Y.~Pang, W.~Wang, F.~E. Tay, W.~Liu, Y.~Tian, and L.~Yuan, ``Masked autoencoders for point cloud self-supervised learning,'' in \emph{European conference on computer vision}.\hskip 1em plus 0.5em minus 0.4em\relax Springer, 2022, pp. 604--621.

\bibitem{zhang2022point}
R.~Zhang, Z.~Guo, P.~Gao, R.~Fang, B.~Zhao, D.~Wang, Y.~Qiao, and H.~Li, ``Point-m2ae: multi-scale masked autoencoders for hierarchical point cloud pre-training,'' \emph{Advances in neural information processing systems}, vol.~35, pp. 27\,061--27\,074, 2022.

\bibitem{zha2024towards}
Y.~Zha, H.~Ji, J.~Li, R.~Li, T.~Dai, B.~Chen, Z.~Wang, and S.-T. Xia, ``Towards compact 3d representations via point feature enhancement masked autoencoders,'' in \emph{Proceedings of the AAAI Conference on Artificial Intelligence}, vol.~38, no.~7, 2024, pp. 6962--6970.

\bibitem{qi2023contrast}
Z.~Qi, R.~Dong, G.~Fan, Z.~Ge, X.~Zhang, K.~Ma, and L.~Yi, ``Contrast with reconstruct: Contrastive 3d representation learning guided by generative pretraining,'' in \emph{International Conference on Machine Learning}.\hskip 1em plus 0.5em minus 0.4em\relax PMLR, 2023, pp. 28\,223--28\,243.

\bibitem{qi2024shapellmuniversal3dobject}
\BIBentryALTinterwordspacing
Z.~Qi, R.~Dong, S.~Zhang, H.~Geng, C.~Han, Z.~Ge, L.~Yi, and K.~Ma, ``Shapellm: Universal 3d object understanding for embodied interaction,'' 2024. [Online]. Available: \url{https://arxiv.org/abs/2402.17766}
\BIBentrySTDinterwordspacing

\bibitem{guo2021pct}
M.-H. Guo, J.-X. Cai, Z.-N. Liu, T.-J. Mu, R.~R. Martin, and S.-M. Hu, ``Pct: Point cloud transformer,'' \emph{Computational Visual Media}, vol.~7, pp. 187--199, 2021.

\bibitem{yang2023ganlmencoderdecoderpretrainingauxiliary}
\BIBentryALTinterwordspacing
J.~Yang, S.~Ma, L.~Dong, S.~Huang, H.~Huang, Y.~Yin, D.~Zhang, L.~Yang, F.~Wei, and Z.~Li, ``Ganlm: Encoder-decoder pre-training with an auxiliary discriminator,'' 2023. [Online]. Available: \url{https://arxiv.org/abs/2212.10218}
\BIBentrySTDinterwordspacing

\bibitem{qi2017pointnetdeeplearningpoint}
\BIBentryALTinterwordspacing
C.~R. Qi, H.~Su, K.~Mo, and L.~J. Guibas, ``Pointnet: Deep learning on point sets for 3d classification and segmentation,'' 2017. [Online]. Available: \url{https://arxiv.org/abs/1612.00593}
\BIBentrySTDinterwordspacing

\bibitem{shapenet2015}
A.~X. Chang, T.~Funkhouser, L.~Guibas, P.~Hanrahan, Q.~Huang, Z.~Li, S.~Savarese, M.~Savva, S.~Song, H.~Su, J.~Xiao, L.~Yi, and F.~Yu, ``{ShapeNet: An Information-Rich 3D Model Repository},'' Stanford University --- Princeton University --- Toyota Technological Institute at Chicago, Tech. Rep. arXiv:1512.03012 [cs.GR], 2015.

\bibitem{wu20153d}
Z.~Wu, S.~Song, A.~Khosla, F.~Yu, L.~Zhang, X.~Tang, and J.~Xiao, ``3d shapenets: A deep representation for volumetric shapes,'' in \emph{Proceedings of the IEEE conference on computer vision and pattern recognition}, 2015, pp. 1912--1920.

\bibitem{liu2019relation}
Y.~Liu, B.~Fan, S.~Xiang, and C.~Pan, ``Relation-shape convolutional neural network for point cloud analysis,'' in \emph{Proceedings of the IEEE/CVF conference on computer vision and pattern recognition}, 2019, pp. 8895--8904.

\bibitem{liu2022masked}
H.~Liu, M.~Cai, and Y.~J. Lee, ``Masked discrimination for self-supervised learning on point clouds,'' in \emph{European Conference on Computer Vision}.\hskip 1em plus 0.5em minus 0.4em\relax Springer, 2022, pp. 657--675.

\bibitem{jiang2023masked}
J.~Jiang, X.~Lu, L.~Zhao, R.~Dazaley, and M.~Wang, ``Masked autoencoders in 3d point cloud representation learning,'' \emph{IEEE Transactions on Multimedia}, 2023.

\bibitem{liu2023pointgame}
Y.~Liu, X.~Yan, Z.~Li, Z.~Chen, Z.~Wei, and M.~Wei, ``Pointgame: Geometrically and adaptively masked autoencoder on point clouds,'' \emph{IEEE Transactions on Geoscience and Remote Sensing}, vol.~61, pp. 1--12, 2023.

\bibitem{chen2023pointgpt}
G.~Chen, M.~Wang, Y.~Yang, K.~Yu, L.~Yuan, and Y.~Yue, ``Pointgpt: Auto-regressively generative pre-training from point clouds,'' \emph{Advances in Neural Information Processing Systems}, vol.~36, pp. 29\,667--29\,679, 2023.

\bibitem{wu2023self}
Y.~Wu, J.~Liu, M.~Gong, P.~Gong, X.~Fan, A.~K. Qin, Q.~Miao, and W.~Ma, ``Self-supervised intra-modal and cross-modal contrastive learning for point cloud understanding,'' \emph{IEEE Transactions on Multimedia}, vol.~26, pp. 1626--1638, 2023.

\bibitem{liu2023inter}
J.~Liu, Y.~Wu, M.~Gong, Z.~Liu, Q.~Miao, and W.~Ma, ``Inter-modal masked autoencoder for self-supervised learning on point clouds,'' \emph{IEEE Transactions on Multimedia}, vol.~26, pp. 3897--3908, 2023.

\bibitem{dong2022autoencoders}
R.~Dong, Z.~Qi, L.~Zhang, J.~Zhang, J.~Sun, Z.~Ge, L.~Yi, and K.~Ma, ``Autoencoders as cross-modal teachers: Can pretrained 2d image transformers help 3d representation learning?'' \emph{arXiv preprint arXiv:2212.08320}, 2022.

\bibitem{zhang2023learning}
R.~Zhang, L.~Wang, Y.~Qiao, P.~Gao, and H.~Li, ``Learning 3d representations from 2d pre-trained models via image-to-point masked autoencoders,'' in \emph{Proceedings of the IEEE/CVF Conference on Computer Vision and Pattern Recognition}, 2023, pp. 21\,769--21\,780.

\bibitem{van2008visualizing}
L.~Van~der Maaten and G.~Hinton, ``Visualizing data using t-sne.'' \emph{Journal of machine learning research}, vol.~9, no.~11, 2008.

\bibitem{mesika2021cloudwalker}
A.~Mesika, Y.~Ben-Shabat, and A.~Tal, ``Cloudwalker: 3d point cloud learning by random walks for shape analysis,'' \emph{arXiv preprint arXiv:2112.01050}, vol.~2, 2021.

\end{thebibliography}

\end{document}